\documentclass[11pt]{article}
\usepackage{acl2015}
\usepackage{times}
\usepackage[normalem]{ulem}
\usepackage[utf8]{inputenc}
\usepackage{graphicx}
\usepackage{amsmath}
\usepackage{amssymb}
\usepackage{tikz}
\usetikzlibrary{arrows,calc,positioning,trees,fit,backgrounds,bayesnet}
\usepackage{tikz-qtree}
\usepackage{tikz-dependency}
\usepackage{xparse}
\usepackage{multirow}
\usepackage{comment}
\usepackage{booktabs}
\usepackage[scientific-notation=true]{siunitx}
\usepackage{url}
\usepackage{paralist}
\usepackage{microtype}
\makeatletter
\newcommand{\@BIBLABEL}{\@emptybiblabel}
\newcommand{\@emptybiblabel}[1]{}
\makeatother
\usepackage{hyperref}
\hypersetup{
    colorlinks=true,
    linkcolor=black,
    citecolor=black,
    filecolor=black,
    urlcolor=black,
}

\DeclareMathOperator*{\argmax}{arg\,max}

\numberwithin{equation}{section} 

\usepackage{relsize}

\usepackage{xspace}
\newcommand{\gf}{syntactic function\xspace}
\newcommand{\gfs}{{\gf}s\xspace}
\newcommand{\Gf}{Syntactic function\xspace}
\newcommand{\Gfs}{{\Gf}s\xspace}

\title{Word Representations, Tree Models and Syntactic Functions}
\author{
 Simon Šuster \\
 University of Groningen\\
 Netherlands \\
 {\tt s.suster@rug.nl} \\ \And
 Gertjan van Noord\\
 University of Groningen\\
 Netherlands\\
 {\tt g.j.m.van.noord@rug.nl}\\
\And
 Ivan Titov\\
 University of Amsterdam \\
 Netherlands\\
 {\tt titov@uva.nl}\\
 }

\begin{document}
\maketitle

\begin{abstract}
Word representations induced from models with discrete latent variables (e.g.\ HMMs) have been shown to be beneficial in many NLP applications.
In this work, we exploit labeled syntactic dependency trees and formalize the induction problem as unsupervised learning of tree-structured hidden Markov models.
Syntactic functions are used as additional observed variables in the model, influencing both transition and emission components.
Such syntactic information can potentially lead to capturing more fine-grain and functional distinctions between words, which, in turn, may be desirable in many NLP applications.
We evaluate the word representations on two tasks -- named entity recognition and semantic frame identification.
We observe  improvements from exploiting syntactic function information in both cases, and the results rivaling those of state-of-the-art representation learning methods.
Additionally, we revisit the relationship between sequential and unlabeled-tree models and find that the advantage of the latter is not self-evident.
\end{abstract}

\section{Introduction}
Word representations have proven to be an indispensable source of features in many NLP systems as they allow better generalization to unseen lexical cases \cite{KooEtAl2008,TurianEtAl2010,TitovKlementiev2012,PassosEtAl2014,BelinkovEtAl2014}.
Roughly speaking,  word representations allow us to capture semantically or otherwise similar lexical items, be it categorically (e.g.\ cluster ids) or in a vectorial way (e.g.\ word embeddings). 
Although the methods for obtaining word representations are diverse, they normally share the well-known distributional hypothesis \cite{Harris1954}, according to which the similarity is established based on occurrence in similar contexts. 
However, word representation methods frequently differ in how they operationalize the definition of context.

Recently, it has been shown that representations using syntactic contexts can be superior to those learned from linear sequences in downstream tasks such as named entity recognition \cite{GraveEtAl2013}, dependency parsing \cite{BansalEtAl2014,SagaeGordon2009} and PP-attachment disambiguation \cite{BelinkovEtAl2014}.
They have also been shown to perform well on datasets for intrinsic evaluation, and to capture a different type of semantic similarity than sequence-based representations \cite{LevyGoldberg2014,SusterVanNoordDepBrown,PadoLapata2007}.

Unlike the recent research in word representation learning, focused heavily on word embeddings from the neural network tradition \cite{CollobertWeston2008,MikolovEtAl2013a,PenningtonEtAl2014}, our work falls into the framework of hidden Markov models (HMMs), drawing on the work of Grave et al.~\shortcite{GraveEtAl2013} and Huang et al.~\shortcite{HuangEtAl2014}.
An attractive property of HMMs is their ability to provide context-sensitive representations, so the same word in two different sentential contexts can be given distinct representations.
In this way, we account for various senses of a word.\footnote{The handling of polysemy and homonymy typically requires extending a model in other frameworks, cf.\ Huang et al.~\shortcite{HuangEtAl2012}, Tian et al.~\shortcite{TianEtAl2014} and Neelakantan et al.~\shortcite{NeelakantanEtAl2014}.}
However, this ability requires inference, which is expensive compared to a simple look-up, so we explore in our experiments word representations that are originally obtained in a context-sensitive way, but are then available for look-up as static representations.


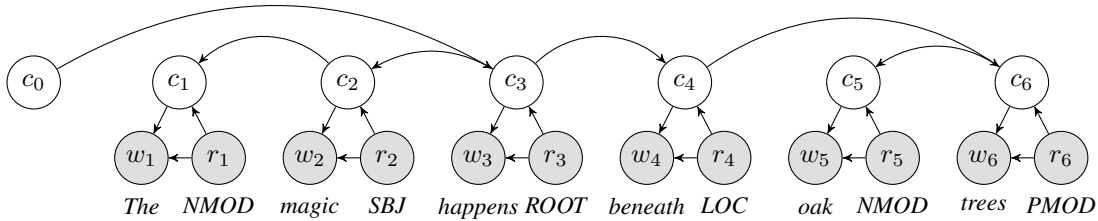
\begin{figure*}[ht]
  \begin{center}
    \begin{tikzpicture}[auto, font=\small]
      \node[obs]                               (w1) {$w_1$};
      \node[obs, right=1.5cm of w1]                               (w2) {$w_2$};
      \node[obs, right=1.5cm of w2]                               (w3) {$w_3$};
      \node[obs, right=1.5cm of w3]                               (w4) {$w_4$};
      \node[obs, right=1.5cm of w4]                               (w5) {$w_5$};
      \node[obs, right=1.5cm of w5]                               (w6) {$w_6$};

      \node[latent] (c1) at (0.5,1) {$c_1$}; 

      \node[latent, right=1.5cm of c1] (c2) {$c_2$}; 
      \node[latent, right=1.5cm of c2] (c3) {$c_3$}; 
      \node[latent, left=1.2cm of c1] (c0) {$c_0$};
      \node[latent, right=1.5cm of c3] (c4) {$c_4$}; 
      \node[latent, right=1.5cm of c4] (c5) {$c_5$}; 
      \node[latent, right=1.5cm of c5] (c6) {$c_6$}; 

      \node[obs, right=0.3cm of w1] (r1) {$r_1$};
      \node[obs, right=0.3cm of w2] (r2) {$r_2$};
      \node[obs, right=0.3cm of w3] (r3) {$r_3$};
      \node[obs, right=0.3cm of w4] (r4) {$r_4$};
      \node[obs, right=0.3cm of w5] (r5) {$r_5$};
      \node[obs, right=0.3cm of w6] (r6) {$r_6$};

      \node[below=0.03cm of w1]                               (The) {\it The};
      \node[below=0.03cm of w2]                               (magic) {\it magic};
      \node[below=0.03cm of w3]                               (happens) {\it happens};
      \node[below=0.03cm of w4]                               (beneath) {\it beneath};
      \node[below=0.03cm of w5]                               (oak) {\it oak};
      \node[below=0.03cm of w6]                               (trees) {\it trees};

      \node[below=0.03cm of r1]                               (NMOD) {\it NMOD};
      \node[below=0.03cm of r2]                               (SBJ) {\it SBJ};
      \node[below=0.03cm of r3]                               (ROOT) {\it ROOT};
      \node[below=0.03cm of r4]                               (LOC) {\it LOC};
      \node[below=0.03cm of r5]                               (NMODD) {\it NMOD};
      \node[below=0.03cm of r6]                               (PMOD) {\it PMOD};




      \path[->, >=stealth', every node/.style={font=\scriptsize}]
      (c1) edge  (w1)
      (c2) edge  (w2)
      (c3) edge  (w3)
      (c4) edge  (w4)
      (c5) edge  (w5)
      (c6) edge  (w6)

      (c3) edge[bend right=30] (c2)
      (c2) edge[bend right=45] (c1)
      (c3) edge[bend left=45] (c4)
      (c4) edge[bend left=35] (c6)
      (c6) edge[bend right=35] (c5)
      (c0) edge[bend left=30] (c3)

      (r1) edge (c1)
      (r2) edge (c2)
      (r3) edge (c3)
      (r4) edge (c4)
      (r5) edge (c5)
      (r6) edge (c6)

      (r1) edge (w1)
      (r2) edge (w2)
      (r3) edge (w3)
      (r4) edge (w4)
      (r5) edge (w5)
      (r6) edge (w6);

\end{tikzpicture}
\caption{Hidden Markov tree model with \gfs, $r$, as additional observed layer.}\label{fig:graphicalmodel}
\end{center}
\end{figure*}

Our method includes two types of observed variables: words and \gfs.
This allows us to address a drawback of learning word representation from unlabeled dependency trees in the context of HMMs (\autoref{sec:synfunc}). 
The motivation for including \gfs comes from the intuition that they act as proxies for semantic roles.
The current research practice is to either discard this type of information (so context words are determined on the syntactic structure alone \cite{GraveEtAl2013}), or include it in a preprocessing step, i.e.\ by attaching syntactic labels to words, as in Levy and Goldberg~\shortcite{LevyGoldberg2014}.

We evaluate the word representations in two structured prediction tasks, named entity recognition (NER) and semantic frame identification.
As our extension builds upon sequential and unlabeled-tree HMMs, we also revisit the basic difference between the two, but are unable to entirely corroborate the alleged advantage of syntactic context for word representations in the NER task.

\section{Why \gfs}\label{sec:synfunc}
A word can typically occur in distinct \gfs. 
Since these account for words in different semantic roles \cite{Bender2013,Levin1993}, the incorporation of the \gf between the word and its parent could give us more precise representations. 
For example,  in ``Carla bought the computer'', the subject and the object represent two different semantic roles, namely the buyer and the goods, respectively.
Along similar lines, Pad\'{o} and Lapata~\shortcite{PadoLapata2007}, Šuster and van Noord~\shortcite{SusterVanNoordDepBrown} and Grave et al.~\shortcite{GraveEtAl2013} argue that it is inaccurate to treat all context words as equal contributors to a word's meaning.

In HMM learning, the parameters obtained from training on unlabeled syntactic structure encode the probabilistic relationship between the hidden states of parent and child, and that between the hidden state and the word.
The tree structure thus only defines the word's context, but is oblivious of the relationship between the words. 
For example, Grave et al.~\shortcite{GraveEtAl2013} acknowledge precisely this limitation of their unlabeled-tree representations by providing as example the hidden state of a verb, which cannot discriminate between left (e.g.\ subject) and right (e.g.\ object) neighbors because of shared transition parameters.
This adversely affects the accuracy of their super-sense tagger for English.
Similarly, Šuster and van Noord~\shortcite{SusterVanNoordDepBrown} show that filtering dependency instances based on \gfs can positively affect the quality of obtained Brown word clusters when measured in a wordnet similarity task.

\section{A tree model with \gfs}
We represent a sentence as a tuple of $K$ words, $\mathbf{w}= (w_1,\ldots,w_K)$, where each $w_k\in\{1,\ldots,|\mathcal{V}|\}$ is an integer representing a word in the vocabulary $\mathcal{V}$.
The goal is to infer a tuple of $K$ states $\mathbf{c} = (c_1,\ldots,c_K)$, where each $c_k \in \{1,\ldots,N\}$ is an integer representing a semantic class of $w_k$, and $N$ is the number of states, which needs to be set prior to training.
Another possibility is to let $w_k$'s representation be a probability distribution over $N$ states.
In this case, we denote $w_k$'s representation as $\mathbf{u}_k \in \mathbb{R}^N$.

As usual in Markovian models, the generation of the sentence can be decomposed into the generation of classes (transitions) and the generation of words (emissions).
The process is defined on a tree, in which a node $c_k$ is generated by its single parent $c_{\pi(k)}$, where $\pi: \{1,\ldots,K\} \mapsto \{0,\ldots,K\}$, with 0 representing the root of the tree (the only node not emitting a word). 
We denote a \gf as $r\in\{r_1,\ldots,r_S\}$, where $S$ is the total number of \gf types produced by the syntactic parser. 
We encode the \gf at position $k$ as $r_k \triangleq r_{w_k \to w_{\pi(k)}}$, i.e.\ the dependency relation between $w_k$ and its parent.

We would like the variable $r$ to modulate the transition and emission processes.
We achieve this by drawing on the Input-output HMM architecture of Bengio and Frasconi~\shortcite{BengioFrasconi1996}, who introduce a sequential model in which an additional sequence of observations called input becomes part of the model, and the model is used as a conditional predictor.
The authors describe the application of their model in speech processing, where the goal is to obtain an accurate predictor of the output phoneme layer from the input acoustic layer. 
Our focus is, in contrast, on representation learning (hidden layer) rather than prediction (output layer). Also, we adapt their sequential topology to trees.

The probability distribution of words and semantic classes is conditional on \gfs and is factorized as:
\vspace{-0.2cm}
\begin{equation}
p(\mathbf{w},\mathbf{c}|\mathbf{r}){=}\prod_{k=1}^K p(w_k|c_k,r_k)p(c_k|c_{\pi(k)},r_k),
\end{equation}

\noindent
where $r_k$ encodes additional information about $w_k$, in our case the \gf of $w_k$ to its parent.
This is represented graphically in fig.\ \ref{fig:graphicalmodel}.

The parameters of the model are stored in column-stochastic transition and emission matrices\footnote{We are abusing the terminology slightly, as these are in fact three-dimensional arrays.}:
\begin{compactitem}
\item[] $\mathbf{T}$, where $T_{ijl} = p(c_k{=}i\ |\ c_{\pi(k)}{=}j,r_k{=}l)$
\item[] $\mathbf{O}$, where $O_{ijl} = p(w_k{=}i\ |\ c_k{=}j,r_k{=}l)$
\end{compactitem}
\vspace{0.2cm}
The number of required parameters for representing the transitions is $O(N^2\, S)$, and for representing the emissions $O(N\, |\mathcal{V}|\, S)$.

Our model satisfies the single-parent constraint and can be applied to proper trees only.
It is in principle possible to extend the base representation for the model by using approximate inference techniques that work on graphs \cite[p. 720]{Murphy2012}, but we do not explore this possibility here.\footnote{This would be relevant for dependency annotation schemes which include secondary edges.}

\begin{figure}[h!]
\begin{center}
\includegraphics[width=\columnwidth]{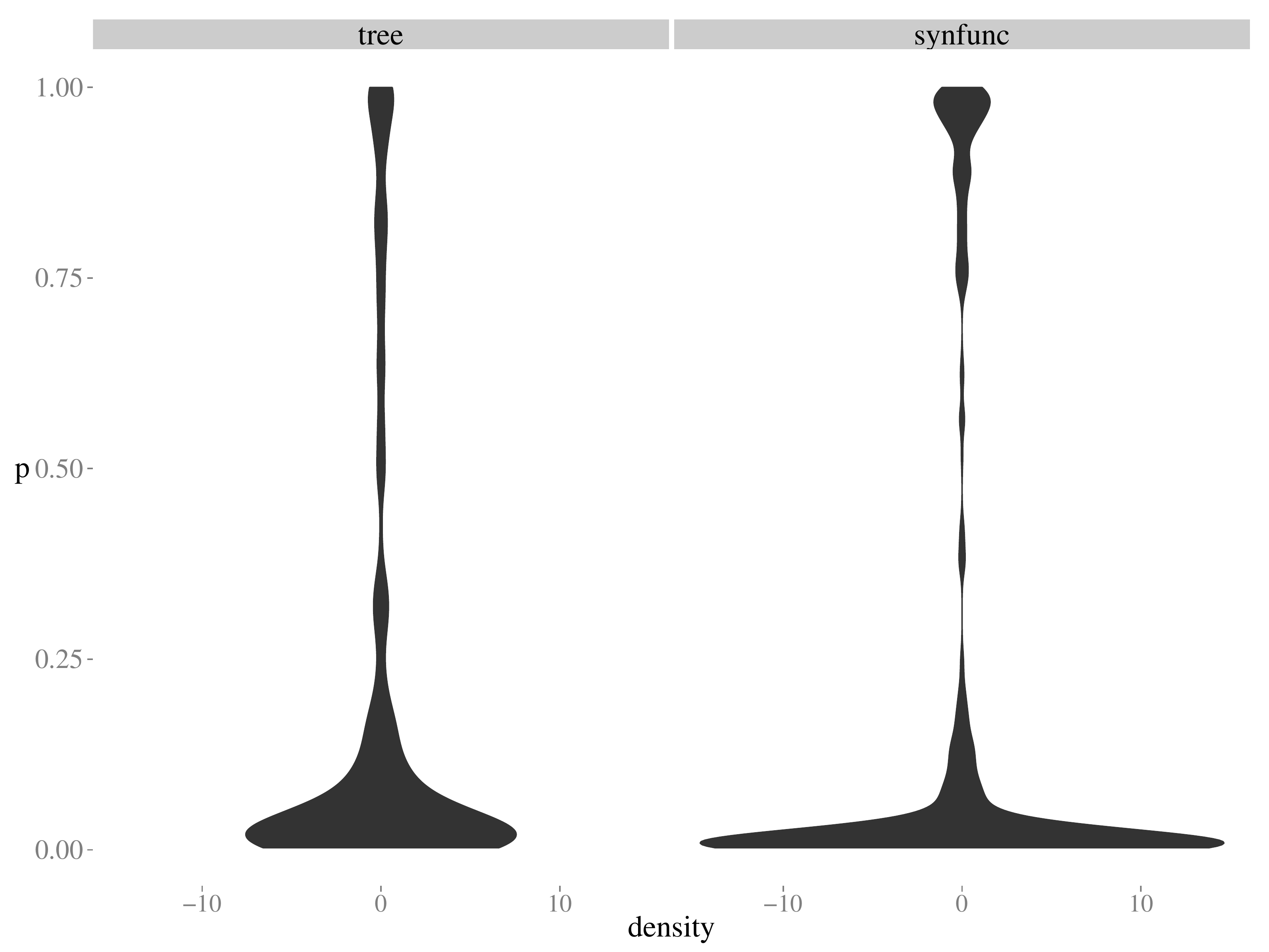}
\end{center}
\caption{The transition probabilities of a tree HMM with syntactic functions ({\em synfunc}) are sparser and have a lower entropy ($5.34$) than those of an unlabeled-tree HMM ({\em tree}; entropy of $5.6$).}
\label{fig:densities}
\end{figure}

As opposed to an unlabeled-tree HMM, our extension can in fact be categorized as an {\it in}homogeneous model since the transition and emission probability distributions change as a function of input, cf.~Bengio~\shortcite{Bengio1999}.
Another comparison concerns the learning of long-term dependencies: since in the Input-output architecture the transition probabilities can change as a function of input at each $k$, they can be more deterministic (have lower entropy) than the transition probabilities of an HMM.
Having the transition parameters closer to zero or one reduces the ambiguity of the next state and allows the context to flow more easily.
A concrete graphical example is given in fig.\ \ref{fig:densities}.

\begin{figure*}
\begin{equation*}
\tau_{ijl} = \sum_{n=1}^N \sum_{k=1}^{K_n} \mathbb{E}\Big[1\{C_k^{(n)}{=}i, C_{\pi(k)}^{(n)}{=}j, R_k^{(n)}{=}l\} \ \Big|\ W^{(n)}{=}\mathbf{w}^{(n)}, R^{(n)}{=}\mathbf{r}^{(n)}\Big]
\end{equation*}
\begin{equation*}
\omega_{ijl} = \sum_{n=1}^N \sum_{k=1}^{K_n} \mathbb{E}\Big[1\{W_k^{(n)}{=}i, C_{\pi(k)}^{(n)}{=}j,  R_k^{(n)}{=}l\}\ \Big|\ W^{(n)}{=}\mathbf{w}^{(n)}, R^{(n)}{=}\mathbf{r}^{(n)}\Big]
\end{equation*}
\caption{Obtaining pseudo-counts, or expected sufficient statistics, in the E-step.}\label{fig:pseudos}
\end{figure*}

\section{Learning and inference}\label{sec:learning}
We train the model with the Expectation-Maximization (EM) algorithm \cite{Baum1972} and use the sum-product message passing for inference on trees \cite{Pearl1988}.
The inference procedure (the estimation of hidden states) is the same as in an unlabeled-tree model, except that it is performed conditionally on $r$.

The parameters $\mathbf{T}$ and $\mathbf{O}$ are estimated with maximum likelihood estimation.
In the E-phase, we obtain pseudo-counts from the existing parameters, as shown in fig.\ \ref{fig:pseudos}.
The M-step then normalizes the transition pseudo-counts (and similarly for emissions):
\begin{equation}
  \label{eq:msteptrans}
  \hat{T}_{ijl} = \frac{\tau_{ijl}}{\sum_{j'}\tau_{ij'l}}
\end{equation}

\subsection{State splitting and merging}\label{sec:statesplitting}
We explore the idea of introducing complexity gradually  in order to alleviate the problem of EM finding a poor solution, which can be particularly severe when the search space is large \cite{PetrovEtAl2006}.
The splitting procedure starts with a small number of states, splits the parameters of each state $s$ into $s_1$ and $s_2$ by cloning $s$ and slightly perturbing.
The model is retrained, and a new split round takes place.
To allow splitting states to various degrees, Petrov et al.\ also merge back those split states which improve the likelihood the least.
Although the merge step is done approximately and does not require new cycles of inference, we find that the extra running time does not justify the sporadic improvements we observe.
We settle therefore on the splitting-only regime.

\subsection{Decoding for HMM-based models}\label{sec:decoding}
Once a model is trained, we can search for the most probable states given observed data by using the max-product message passing  (\textsc{Max-Product}, a generalization of Viterbi) for efficient decoding on trees: 
 $\mathbf{\hat{c}} = \argmax_{\mathbf{c}} p(C=\mathbf{c}\ |\ W=\mathbf{w},R=\mathbf{r}).$\label{eq:maxproduct}

 We have also tried posterior (or minimum risk) decoding \cite{LemberAndKoloydenko2014,GanchevEtAl2008}, but without consistent improvements.

The search for the best states can be avoided by taking the posterior state distribution $\mathbf{u}_k$ over $N$ hidden states \cite{NepalYates2014,GraveEtAl2014}:
  $u_c^{(k)} = \mathbb{E}\big[1\{C_k=c\}\ \big|\ W=\mathbf{w},R=\mathbf{r}\big].$\label{eq:posteriortoken}
We call this vectorial representation \textsc{Post-Token}.

In both cases, inference is performed on a concrete sentence, thus providing a context-sensitive representation.
We find in our experiments that \textsc{Post-Token} consistently outperforms \textsc{Max-Product} due to its ability to carry more information and uncertainty.
This can then be exploited by the downstream task predictor.

One disadvantage of obtaining context-sensitive representations is the relatively costly inference.
The inference and decoding are also sometimes not applicable, such as in information retrieval, where the entire sentence is usually not given \cite{HuangEtAl2011}.
A trade-off between full context sensitivity and efficiency can be achieved by considering a static representation (\textsc{Post-Type}).
It is obtained in a context-insensitive way \cite{HuangEtAl2011,GraveEtAl2014} by averaging posterior state distributions (context-sensitive) of all occurrences of a word type $\tilde{w}$ from a large corpus $\mathcal{U}$:
\begin{equation}
  \mathbf{v}^{(\tilde{w})} = \frac{1}{Z_{\tilde{w}}} \sum_{i \in \mathcal{U}:w_i=\tilde{w}} \mathbf{u}^{(i)}.
\end{equation}\label{eq:posteriortype}

\begin{figure}[h!]
\begin{center}
\includegraphics[width=\columnwidth]{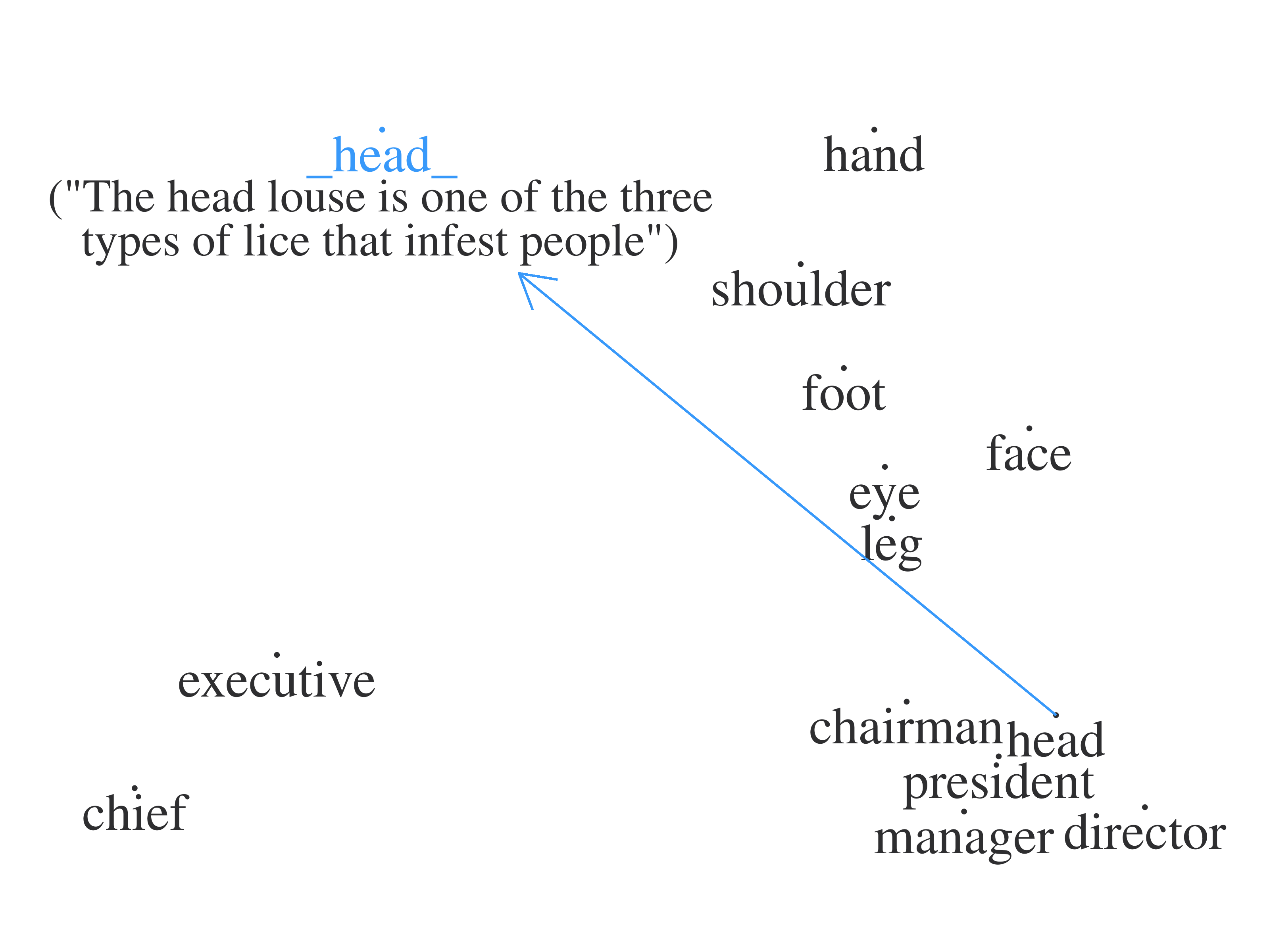} \end{center}
\caption{Representations obtained with our model with \gfs. 
All are static {\sc Post-Type} representations, except ``\_head\_'', which is obtained with {\sc Post-Token} from the concrete sentence.}\label{fig:words}
\end{figure}

In fig.\ \ref{fig:words}, we give a graphical example of the word representations learned with our model (\autoref{sec:hmmdetails}), obtained either with the {\sc Post-Token} or the {\sc Post-Type}.
To visualize the representations, we apply multidimensional scaling.\footnote{\url{https://github.com/scikit-learn/scikit-learn}}
The model clearly separates between management positions and parts of body, and interestingly, puts ``head'' closer to management positions, which can be explained by the business and economic nature of the Bllip corpus.
The words ``chief'' and ``executive'' are located together, yet isolated from others, possibly because of their strong tendency to precede nouns. 
The arrow on the plot indicates the shift in the meaning when a {\sc Post-Token} representation is obtained for ``head'' (part-of-body) within a sentence.
 
Despite the advantage of {\sc Post-Token} to account for word senses, we observe that {\sc Post-Type} performs better in almost all experiments.
A likely explanation is that averaging increases the generalizability of representations. 
For the concrete tasks in which we apply the word representations, the increased robustness is simply more important than context sensitivity.
Also, {\sc Post-Type} might be less sensitive to parsing errors during test time.

\section{Empirical study}
\subsection{Parameters and setup}
We observe faster convergence times with online EM, which updates the parameters more frequently.
Specifically, we use the mini-batch step-wise EM \cite{LiangKlein2009,CappeMoulines2009}, and determine the hyper-parameters on the held-out dataset of 10,000 sentences to maximize the log-likelihood.
We find out that higher values for the step-wise reduction power $\alpha$ and the mini-batch size lead to better overall log-likelihood, but with a somewhat negative effect on the convergence speed.
We finally settle on $\alpha=\{0.6,0.7,0.8,0.9,\underline{1}\}$ and mini-batch size of $\{1,10,100,\underline{1000}\}$ sentences.
We find that a couple of iterations over the entire dataset is sufficient to obtain good parameters, cf.\ Klein~\shortcite{Klein2005}.

\noindent
{\bf Initialization. }
Since the EM algorithm in our setting only finds a local optimum of the log-likelihood, the initialization of model parameters can have a major impact on the final outcome.
We initialize the emission matrices with Brown clusters by first assigning random values between 0 and 1 to the matrix elements, and then multiplying those elements that represent words in a cluster by a factor of $f\in{\{10,100,\underline{1K},10K\}}$.
Finally, we normalize the matrices.
This technique incorporates a strong bias towards word-class emissions that exist (deterministically) in Brown clusters.
The transition parameters are simply set to random numbers sampled from the uniform distribution between 0 and 1, and finally normalized.

\noindent
{\bf Approximate inference. }
Following Grave et al.~\shortcite{GraveEtAl2013} and Pal et al.~\shortcite{PalEtAl2006}, we approximate the belief vectors during inference,\footnote{In a bottom-up pass, a belief vector represents the local evidence by multiplying the messages received from the children of a node, as well as the emission probability at that node.} which speeds up learning and works as regularization.
We use the $k$-best projection method, in which only $k$-largest coefficients (in our case $k=\frac{1}{8}N$) are kept.

\subsection{Data for obtaining word representations}
{\bf English.} 
We use the 43M-word Bllip corpus \cite{CharniakEtAl2000} of WSJ texts, from which we remove the sentences of the PTB and those whose length is $\leq 4$ or $\geq 40$.
We use the MST dependency parser \cite{McDonaldPereira2006} for English and build a projective, second order model, trained on sections 2--21 of the Penn Treebank WSJ (PTB).
Prior to that, the PTB was patched with NP bracketing rules \cite{VadasCurran2007} and converted to dependencies with LTH \cite{JohanssonNugues2007}.
The parser achieves the unlabeled/labeled accuracy of 91.5/85.22 on section 23 of the PTB without retagging the POS.
For POS-tagging the Bllip corpus and the evaluation datasets, we use the Citar tagger \footnote{\url{http://github.com/danieldk/citar}}.
After parsing, we replace the words occurring less than 40 times with a special symbol to model OOV words.
This results in the vocabulary size of $~$27,000 words.

\noindent
{\bf Dutch.} We first produce a random sample of 2.5M sentences from the SoNaR corpus\footnote{\small\url{http://lands.let.ru.nl/projects/SoNaR}}, then follow the same preprocessing steps as for English. 
We parse the corpus with Alpino \cite{vanNoord2006}, an HPSG parser with a maxent disambiguation component.
In contrast with English, for which we use word forms, we keep here the root forms produced by the parser's lexical analyzer.
The resulting vocabulary size is $~$25,000 words.
The analyses produced by the parser represent multiple parents to facilitate the treatment of  wh-clauses, coordination and passivization. 
Since our method expects proper trees, we convert the parser output to CoNLL format.\footnote{\url{http://www.let.rug.nl/bplank/alpino2conll/}}

\subsection{Evaluation tasks}\label{sec:eval}
{\bf Named entity recognition.} We use the standard CoNLL-2002 shared task dataset for Dutch and CoNLL-2003 dataset for English.
 We also include the out-of-domain MUC-7 testset, preprocessed according to Turian et al.~\shortcite{TurianEtAl2010}. 
We refer the reader to Ratinov and Roth~\shortcite{RatinovAndRoth2009} for a detailed description of the NER classification problem.

Just like Turian et al.~\shortcite{TurianEtAl2010}, we use the averaged structured perceptron \cite{Collins2002} with Viterbi as the base for our NER system.\footnote{\url{http://github.com/LxMLS/lxmls-toolkit}}
 The classifier is trained for a fixed number of iterations, and uses these baseline features:
\begin{compactitem}
\item $w_k$ information: is-alphanumeric, all-digits, all-capitalized, is-capitalized, is-hyphenated;
\item prefixes and suffixes of $w_k$;
\item word window: $w_k, w_{k\pm1}, w_{k\pm2}$\label{wordwindow};
\item capitalization pattern in the word window.
\end{compactitem}

We construct $N$ real-valued features for a word vector of dimensionality $N$, and a simple indicator feature for a categorical word representation.

\noindent
{\bf Semantic frame identification.} Frame-semantic parsing is the task of identifying \begin{inparaenum}[\itshape a\upshape)]
\item \label{itm:frameid} semantic frames  of predicates in a sentence (given {\em target} words evoking frames), and 
\item frame arguments participating in these events \cite{Fillmore1982,DasEtAl2014}. 
\end{inparaenum}
Compared to NER, in which the classification decisions apply to a relatively small set of words, the problem of semantic frame identification concerns making predictions for a broader set of words (verbs, nouns, adjectives, sometimes prepositions).

\begin{figure}[h!]
\includegraphics[width=\columnwidth]{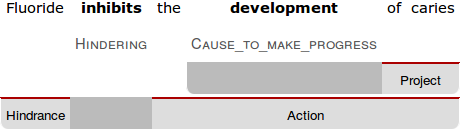}
\caption{A parse with {\sc Hindering} and {\sc Cause\_to\_make\_progress} frames with respective arguments.}
\label{fig:fspex1}
\end{figure}
We use the Semafor parser \cite{DasEtAl2014} consisting of two log-linear components trained with gradient-based techniques.
The parser is trained and tested on the FrameNet 1.5 full-text annotations.
Our test set consists of the same 23 documents as in Hermann et al.~\shortcite{HermannEtAl2014}.
We investigate the effect of word representation features on the frame identification component. 
We measure Semafor's performance on gold-standard targets, and report the accuracy on {\em exact matches}, as well as on {\em partial matches}. 
The latter give partial credit to identified related frames.
We use and modify the publicly available implementation at \url{http://github.com/sammthomson/semafor}.

Our baseline features for a target $w_k$ include:
\begin{compactitem}
\item $w_k$ and $w_{\pi(k)}$ (if the parent is a preposition, the grandparent is taken by collapsing the dependency),
\item their lemmas and POS tags,
\item \gfs between:
  \begin{compactitem}
  \item $w_k$ and its children,
  \item $w_k$ and $w_{\pi(k)}$ (added by ourselves),
  \item $w_{\pi(k)}$ and its parent $w_{\pi(\pi(k))}$.
  \end{compactitem}
\end{compactitem}


\subsection{Baseline word representations}
We test our model, which we call {\sc Synfunc-Hmm}, against the following baselines:
\begin{compactitem}
\item {\sc Baseline}: no word representation features
\item {\sc Hmm}: a sequential HMM
\item {\sc Tree-Hmm}: a tree HMM
\end{compactitem}

\noindent
We induce other representations for comparison:
\begin{compactitem}
\item {\sc Brown}: Brown clusters
\item {\sc Dep-Brown}: dependency Brown clusters
\item {\sc Skip-Gram}: Skip-Gram word embeddings
\end{compactitem}

\subsection{Preparing word representations}\label{sec:hmmdetails}
\begin{table*}[ht!]
\setlength\tabcolsep{4pt}
\hfill{}
{\footnotesize
\begin{tabular}{l l l l l l l l l l}
                 & \multicolumn{3}{c}{English} & \multicolumn{3}{c}{Dutch} & \multicolumn{3}{c}{MUC test set}                                                    \\
\cmidrule(lr){2-4}
\cmidrule(lr){5-7}
\cmidrule(lr){8-10}
Model             & P & R           & F-1         & P              & R           & F-1         & F-1                   & $\text{F-1}_{\text{type}}$ & $\text{F-1}_{\text{unlab}}$           \\
\cmidrule(lr){1-10}
{\sc Baseline}        & 80.12       & 77.30       & 78.69          & 75.36       & 70.92       & 73.07                 & 65.44                      & 87.04          & 96.25                \\
{\sc Hmm}             & {\bf 81.49} & 78.90       & {\bf 80.17}    & 77.61       & 73.97       & 75.74                 & 70.20                      & 87.66          & 96.50                \\
{\sc Tree-Hmm}        & 80.49       & 78.10       & 79.28          & 77.41       & 73.48       & 75.40                 & 65.67                      & 86.99          & 96.53                \\
{\sc Synfunc-Hmm}     & 80.65       & 78.90       & 79.76 (+.48)   & {\bf 78.54} & {\bf 75.23} & {\bf 76.85} (+1.45)   & 66.49 (+.82)               & 86.93 (-.06)   & {\bf 96.69} (+.16) \\
{\sc Brown}           & 80.15       & 77.28       & 78.70          & 77.88       & 71.73       & 74.68                 & 68.85                      & 87.72          & 96.67                \\
{\sc Dep-Brown}       & 78.80       & 75.73       & 77.23          & 77.50       & 73.66       & 75.53                 & 68.31                      & 87.44          & 96.47                \\
{\sc Skip-Gram}       & 80.80       & {\bf 78.98} & 79.88          & 76.02       & 71.28       & 73.57                 & {\bf 72.42}                & {\bf 88.94}    & {\bf 96.69}          \\
\end{tabular}}

\hfill{}
\caption{NER results (precision, recall and F-score) on English and Dutch test sets.
Best result per column in bold.
The score increase reported in parentheses is in comparison to {\sc Tree-Hmm}.
F-1$_{\text{type}}$ is the F-score measured per word type, and F-1$_{\text{unlab}}$ is the F-score measured per word type, ignoring labels.}\label{tab:neren}
\end{table*}

\noindent
{\bf Brown clusters.} 
Brown clusters \cite{BrownEtAl1992} are known to be effective and robust when compared, for example, to word embeddings \cite{BansalEtAl2014,PassosEtAl2014,NepalYates2014,QuEtAl2015}.
The method can be seen as a special case of a HMM in which word emissions are deterministic, i.e.\ a word belongs to at most one semantic class.
Recently, an extension has been proposed on the basis of a dependency language model \cite{SusterVanNoordDepBrown}. 
We use the publicly available implementations.\footnote{\url{http://github.com/percyliang/brown-cluster},\\\url{http://github.com/rug-compling/dep-brown-cluster}}

Following other work on English \cite{KooEtAl2008,NepalYates2014}, we add both coarse- and fine-grained clusters as features by using prefixes of length 4, 6, 10 and 20 in addition to the complete binary tree path.
For Dutch, coarser-grained clusters do not yield any improvement.
Brown features are included in a window around the target word, just as the NER word features.
When adding cluster features to the frame-semantic parser, we transform the cluster identifiers to one-hot vectors, which gives a small improvement over the use of indicator features.

\noindent
{\bf HMM-based models.} The $N$-dimensional representations obtained from HMMs and their variants are included as $N$ distinct continuous features.
In the NER task, word representations are included at $w_k$ and $w_{k+1}$ for Dutch and at $w_k$ for English, which we determined on the development set.
We investigate state space sizes of $64$, $128$ and $256$ and finally choose $N{=}128$ as a reasonable trade-off between training time and quality.
We use the same dimensionality for other word representation models in this paper.

We observe that by constraining {\sc Synfunc-Hmm} to use only the $k$ most frequent \gfs and to treat the remaining ones as a single special \gf, we obtain better results in our evaluation tasks.
This is because for a model with all $S$ \gfs produced by the parser, there is less learning evidence for more infrequent \gfs.
We explore the effect of keeping up to five most frequent \gfs, ignoring functional ones such as punctuation and determiner.\footnote{We define the list of function-marker \gfs following Goldberg and Orwant~\shortcite{GoldbergOrwant2013}.}
The final selection is shown in table \ref{tab:rels}.

\begin{table}[ht!]
\resizebox{\columnwidth}{!}{
\begin{tabular}{l l}
English               & Dutch \\
\midrule
nmod (nominal modifier)      & mod (modifier) \\
pmod (prepositional modifier) & su (subject) \\
sub (subject)               & obj1 (direct object) \\
                      & cnj (conjunction) \\
                      & mwp (multiword unit) \\
\end{tabular}}
\caption{\Gfs in {\sc Synfunc-Hmm} for English (produced by the MST parser) and Dutch (produced by Alpino).  }\label{tab:rels}
\end{table}


For NER experiments, the representations from all HMM models are obtained with three different ``decoding'' methods (\autoref{sec:decoding}).
Since {\sc Post-Type} is performing best overall, we only report the results for this method in the evaluation.\footnote{While exploring the constraint on the number of \gfs, we do find that {\sc Post-Token} outperforms {\sc Post-Type} in some sets of \gfs, but not in the final, best-performing selection.}

\noindent
{\bf Word embeddings.}
We use the Skip-Gram model presented in Mikolov et al.~\shortcite{MikolovEtAl2013a} (\url{https://code.google.com/p/word2vec/}), trained with negative sampling \cite{MikolovEtAl2013b}.
The training seeks to maximize the dot product between word-context pairs encountered in the training corpus and minimize the dot product between those pairs in which the context word is randomly sampled.
We set both the number of negative examples and the size of the context window to $5$, the down-sampling threshold to \num{1e-4}, and the number of iterations to $15$.

\subsection{NER results}
The results for all testsets are shown in table \ref{tab:neren}.
For English, all HMM-based models improve the baseline, with the sequential {\sc Hmm} achieving the highest F-score. 
Our {\sc Synfunc-Hmm} performs on a par with {\sc Skip-Gram}. 
It outperforms the unlabeled-tree model, indicating that the added observations are useful and correctly incorporated. 
Brown clusters do not exceed the {\sc Baseline} score.\footnote{However, after additional experiments we observe that the cluster features do improve over the baseline score when the number of clusters is increased.}
Testing for significance with a bootstrap method \cite{SogaardEtAl2014}, we find out that only {\sc HMM} improves significantly at $p<0.01$ on macro-F1 over {\sc Baseline}, while {\sc Skipgram} and {\sc Synfunc-Hmm} show significant improvements only for the location entity type.

The general trend for Dutch is somewhat different. 
Most notably, all word representations contribute much more effectively to the overall classification performance compared to English. 
The best-scoring model, our {\sc Synfunc-Hmm}, improves over the baseline significantly by as much as about 3.8 points.
Part of the reason {\sc Synfunc-Hmm} works so well in this case is that it can make use of the informative ``mwp'' \gf between the parts of a multiword unit. 
Similarly as for English, the unlabeled-tree HMM performs slightly worse than the sequential {\sc Hmm}. 
The cluster features are more valuable here than in English, and we also observe a 0.7-point advantage by using dependency Brown clusters over the standard, bigram Brown clusters. 
The {\sc Skip-Gram} model does not perform as well as in English, which might indicate that the hyper-parameters would need fine-tuning specific to Dutch.

On the out-of-domain MUC dataset, tree-based representations appear to perform poorly, whereas the highest score is achieved by the {\sc Skip-Gram} method. 
Unfortunately, it is difficult to generalize from these $\text{F-1}$ results alone.
Concretely, the dataset contains 3,518 named entities, and the {\sc Skip-Gram} method makes 258 correct predictions more than {\sc Tree-Hmm}. 
However, because the MUC dataset covers the narrow topic of missile-launch scenarios, the system gets badly penalized if a mistake is made repeatedly for a certain named entity. 
For example, only the entity ``NASA'' occurs 103 times, most of which are wrongly classified by the {\sc Tree-Hmm} system, but correctly by {\sc Skip-Gram}.
The overall performance may therefore hinge on a limited number of frequently occurring entities.
A workaround is to evaluate per {\it entity type} --- calculate the F-score for each entity, then average over all entity types.
The results for this evaluation scenario are reported as $\text{F-1}_{\text{type}}$. 
{\sc Skip-Gram} still performs best, but the difference to other models is smaller.
Finally, we also report $\text{F-1}_{\text{unlab}}$, calculated as $\text{F-1}_{\text{type}}$ but ignoring the actual entity label.
So, if a named-entity token is recognized as such, we count it as correct prediction ignoring the entity label type, similarly as done by Ratinov and Roth~\shortcite{RatinovAndRoth2009}.
Since {\sc Synfunc-Hmm} performs better here, we can conclude that it is more effective at identifying entities rather than at labeling them.

The fact that we observe different tendencies for English and Dutch can be attributed to an interplay of factors, such as language differences \cite{Bender2011}, differently-performing syntactic parsers, and differences specific to the evaluation datasets.
We briefly discuss the first possibility. 
It is clear from table \ref{tab:neren} that all syntax-based models ({\sc Dep-Brown}, {\sc Tree-Hmm}, {\sc Synfunc-Hmm}) generally benefit Dutch more than English. 
We hypothesize that since the word order in Dutch is generally less fixed than in English,\footnote{For instance, it is unusual for the direct object in English to precede the verb, but quite common in Dutch.} a sequence-based model for Dutch cannot capture selectional preferences that successfully, i.e.\ there is more interchanging of semantically diverse words in a small word window.
This then makes the difference in performance between sequential and tree models more apparent for Dutch.

\subsection{Semantic frame identification results}
The results are shown in table \ref{tab:fspframeid}.
The best score is obtained by the Skip-Gram embeddings, however, the difference to other models outperforming the baseline is small. 
For example, {\sc Skip-Gram} correctly identifies only two cases more than {\sc Dep-Brown}, out of 3681 correctly disambiguated frames.

The {\sc Synfunc-Hmm} model outperforms all other HMM models in this task.
The differences are larger when scoring partial matches.\footnote{On exact matches, only {\sc Dep-Brown} and {\sc Brown} significantly outperform the baseline with the $p<0.05$.
On partial matches, {\sc Dep-Brown}, {\sc Brown}, {\sc Skip-Gram} and {\sc Synfunc-Hmm} all outperform the baseline significantly.
{\sc Synfunc-Hmm} performs significantly better than {\sc Tree-Hmm} on partial matches, whereas the difference between {\sc Skip-Gram} and {\sc Synfunc-Hmm} is not significant. The significance tests were run using paired permutation.}

\begin{table}[ht!]
\resizebox{\columnwidth}{!}{
\begin{tabular}{l p{0.5cm} l p{0.5cm} l}
  Model           & Exact       &                   & Partial     &                   \\
\cmidrule(lr){1-5}
{\sc Baseline}    & 82.70       &                   & 90.44       &                   \\ 
{\sc Hmm}         & 82.20       &                   & 90.20       &                   \\ 
{\sc Tree-Hmm}    & 82.89       &                   & 90.59       &                   \\  
{\sc Synfunc-Hmm} & 82.95       & (+0.06) & 90.80       & (+0.21) \\
{\sc Brown}       & 83.15       &                   & 90.74       &                   \\
{\sc Dep-Brown}   & 83.15       &                   & 90.76       &                   \\
{\sc Skip-Gram}   & {\bf 83.19} &                   & {\bf 90.91} &                   \\
\end{tabular}}
\caption{Frame identification accuracy.
Score increase in parentheses is relative to {\sc Tree-Hmm}.}\label{tab:fspframeid}
\end{table}

\subsection{Further discussion}
We can conclude from the NER experiments that {\em unlabeled} syntactic trees do not in general provide a better structure for defining the contexts compared to plain sequences.
The only exception is the case of dependency Brown clustering for Dutch.
Comparing our results to those of Grave et al.~\shortcite{GraveEtAl2013}, we therefore cannot confirm the same advantage when using unlabeled-tree representations.
In semantic frame identification, however, the unlabeled-tree representations do compare more favorably to sequential representations.

Our extension with \gfs outperforms the baseline and other HMM-based representations in practically all experiments.
It also outperforms all other word representations in Dutch NER.
The advantage comes from discriminating between the types of contexts, for example between a modifier and a subject, which is impossible in sequential or unlabeled-tree HMM architectures.
The results for English are comparable to those of the state-of-the-art representation methods.


\section{Related work}

HMMs have been used successfully for learning word representations already before, see Huang et al.~\shortcite{HuangEtAl2014} for an overview, with an emphasis on investigating domain adaptability. 
Models with more complex architecture have been proposed, such as a factorial HMM \cite{NepalYates2014}, trained using approximate variational inference and applied to POS tagging and chunking.
Recently, semantic compositionality of HMM-based representations based on the framework of distributional semantics has been investigated by Grave et al.~\shortcite{GraveEtAl2014}. 

There is a long tradition of unsupervised training of HMMs for POS tagging \cite{Kupiec1992,Merialdo1994}, with more recent work on incorporating bias by favoring sparse posterior distributions within the posterior regularization framework \cite{GracaEtAl2007}, and for example on auto-supervised refinement of HMMs \cite{GarretteAndBaldridge2012}. 
It would be interesting to see how well these techniques could be applied to word representation learning methods like ours.

The extension of HMMs to dependency trees for the purpose of word representation learning was first proposed by Grave et al.\ \shortcite{GraveEtAl2013}. 
Although our baseline HMM methods, {\sc Hmm} and {\sc Tree-Hmm}, conceptually follow the models of Grave et al., there are still several practical differences. 
One source of differences is in the precise steps taken when performing Brown initialization, state splitting, and also approximation of belief vectors during inference.
Another source involves the evaluation setting. 
Their NER classifier uses only a single feature, and the inclusion of Brown clusters does not make use of the clustering hierarchy. 
In this respect, our experimental setting is more similar to Turian et al.~\shortcite{TurianEtAl2010}. 
Another practical difference is that Grave et al.\ concatenate words with POS-tags to construct the input text, whereas we use tokens (English) or word roots (Dutch).

The incorporation of word representations into semantic frame identification has been explored in Hermann et al.~\shortcite{HermannEtAl2014}.
They perform a projection of generic word embeddings for context words to a low-dimensional representation, which also learns an embedding for each frame label.
The method selects the frame closest to the low-dimensional representation obtained through mapping of the input embeddings.
Their approach differs from ours in that they induce new representations that are tied to a specific application, whereas we aim to obtain linguistically enhanced word representations that can be subsequently used in a variety of tasks.
In our case, the word representations are thus included as additional features in the log-linear model.
The inclusion accounts for \gfs between the target and its context words.
Although Hermann et al.\ also use \gfs, they are used to position the general word embeddings within a single input context embedding.
Unfortunately, we are unable to directly compare our results with theirs as their parser implementation is proprietary.
The accuracy of our baseline system on the test set is $0.27$ percent lower in the exact matching regime and $0.07$ lower in the partial matching regime compared to the baseline implementation \cite{DasEtAl2014} they used.\footnote{Among other implementation differences, they introduce a variable capturing lexico-semantic relations from WordNet.}

The topic of context type (syntactic vs.\ linear) has been abundantly treated in distributional semantics \cite{Lin1998Syntactic,BaroniLenci2010,vanDeCruys2010} and elsewhere \cite{BoydGraberBlei2008,TjongkimsangAndHofmann2009}.


\section{Conclusion}

We have proposed an extension of a tree HMM  with \gfs.
The obtained word representations achieve better performance than those from the unlabeled-tree model.
Our results also show that simply preferring an unlabeled-tree model over a sequential model does not always lead to an improvement.
An important direction for future work is to investigate how discriminating between context types can lead to more accurate models in other frameworks.
The code for obtaining HMM-based representations described in this paper is freely available at \url{http://github.com/rug-compling/hmm-reps}.


\section*{Acknowledgements}
We would like to thank Edouard Grave and Sam Thomson for valuable discussion and suggestions.
\bibliographystyle{acl}
\bibliography{0latexLiterature.bib}

\end{document}